\documentclass[letterpaper,twocolumn,10pt]{article}
\usepackage{usenix2019_v3}
\usepackage{lipsum}
\usepackage{graphicx}
\usepackage{titlesec}
\titlespacing*{\section}{0pt}{1.1\baselineskip}{\baselineskip}

\usepackage{tikz}
\usepackage{amsmath}
\usepackage{enumitem}

\usepackage{filecontents}
\usepackage{algorithm, algorithmicx, algpseudocode}
\usepackage{verbatim}
\usepackage[font={small,it}]{caption}
\usepackage{caption}
\usepackage[caption=false]{subfig}
\usepackage{mathtools}

\newcommand{\arjun}[1]{}
\newcommand{\adarsh}[1]{}
\newcommand{\shivaram}[1]{}
\newcommand{\aditya}[1]{}
\renewcommand{\arjun}[1]{{\color{red}{\bf [AB: #1}]}}
\renewcommand{\adarsh}[1]{{\color{brown}{\bf [AK: #1}]}}
\renewcommand{\shivaram}[1]{{\color{blue}{\bf [SV: #1}]}}
\renewcommand{\aditya}[1]{{\color{green}{\bf [AA: #1}]}}

\begin{document}

\date{}

\title{\Large \bf Accelerating Deep Learning Inference via Freezing }

\author{
{\rm Adarsh Kumar \hspace{5mm} \rm Arjun Balasubramanian  \hspace{5mm} \rm Shivaram Venkataraman \hspace{5mm} \rm Aditya Akella}\\ \\
University of Wisconsin - Madison
} 
\maketitle
\pagestyle{empty}

\begin{abstract}
\label{sec:design}

Over the last few years, Deep Neural Networks (DNNs) have become ubiquitous owing to their high accuracy on real-world tasks. However, this increase in accuracy comes at the cost of computationally expensive models leading to higher prediction latencies. Prior efforts to reduce this latency such as quantization, model distillation, and any-time prediction models typically trade-off accuracy for performance. In this work, we observe that caching intermediate layer outputs can help us avoid running all the layers of a DNN for a sizeable fraction of inference requests. We find that this can potentially reduce the number of effective layers by half for 91.58\% of CIFAR-10 requests run on ResNet-18. We present Freeze Inference, a system that introduces \emph{approximate caching} at each intermediate layer and we discuss techniques to reduce the cache size and improve the cache hit rate. Finally, we discuss some of the open research challenges in realizing such a design.

\end{abstract}

\section{Introduction}
\label{sec:design}
The field of artificial intelligence (AI) has made rapid strides over the past few years largely due to the progress in Deep Neural Networks (DNNs). DNNs have surpassed human-level accuracy on tasks ranging from speech recognition\cite{toward-human-parity-conversational-speech-recognition}, image classification\cite{NIPS2012_4824,DBLP:journals/corr/SimonyanZ14a, DBLP:journals/corr/LinCY13, 43022, DBLP:journals/corr/SzegedyVISW15,Resnet} to machine translation~\cite{hassan2018achieving}. However, this gain in accuracy has come with models becoming deeper leading to increased computational requirements. For example, in object classification, the top-5 classification accuracy has increased from 71\% in 2012 to 97\% in 2015 on the ImageNet dataset, while the models have become 20$\times$ more computationally expensive. This increase in computation also leads to longer latencies during prediction or inference where low user response time is paramount. To efficiently serve these models, there is a need to reduce the overall computation needed for inference, without trading off accuracy.
 
 
There have been several efforts to reduce the computational complexity of DNNs to improve model serving. A number of previous efforts have proposed compressing the model using techniques such as quantization ~\cite{DBLP:journals/corr/CourbariauxB16,DBLP:journals/corr/CaiHSV17} or model distillation ~\cite{hinton2015distilling}, but such techniques typically hurt accuracy. On the other hand, ensemble methods~\cite{mcdnnPaper} or any-time prediction models~\cite{DBLP:journals/corr/HuangCLWMW17} aim to provide a better trade-off between accuracy and latency by building models of varying complexity. However, this either requires re-training using custom model architectures or training a number of models ahead of time. Systems such as Clipper~\cite{crankshaw2017clipper} improve serving by batching queries and optimizing their execution within a batch. These techniques typically improve throughput and making inference latency-aware. Finally, PRETZEL\cite{pretzel} improves latencies using multi-model optimizations but does not focus on reducing compute for a single given model.



In this work, we introduce \emph{caching} as a technique to reduce the prediction latency of DNNs. Caches in general are used to improve the latency of Web requests by storing the output of previous requests. Previous works \cite{crankshaw2017clipper} have used caching at the input layer  to improve the prediction latency, but they consider the DNN as a black box. Thus, in the event of a cache miss at the input layer, these systems have to run all the layers of the DNN to obtain a prediction. Instead the question we ask is: \emph{Can we design caching such that we can avoid the need to run all the layers of a DNN for every input request?}

To this end, we propose augmenting DNNs with a cache at each layer, where the cache holds a succinct representation of intermediate layer outputs and their relation to the final classification. The rationale behind this is that each layer of the DNN tries to normalize the variations in the input that do not correlate with the output, and by doing so, tries to learn an embedding space where similar data points are closer to each other. Thus, even when we do not get a "cache hit" for the input, we could get a "cache hit" at an intermediate layer. For example, with an object classification model, the background, brightness, contrast, etc of the input image do not correlate with the output class. The model will normalize these variations in the image, layer by layer. Thus, we can expect images of the same object with different backgrounds to be embedded closer in the projection space after some DNN layers have been evaluated.

Maintaining a cache for intermediate layers comes with its own challenges. First, as the intermediate layer outputs are float tensors in a high dimensional space, the probability of exact match is low, leading to a low cache hit rate. Second, such a cache would require a large amount of memory owing to the high dimensionality of tensors and the size of training data. Finally, cache look-up time is poor for large caches that cannot fit in fast memories. 

We introduce Freeze Inference, a system which augments DNNs with intermediate layer caches to reduce the prediction latency and addresses the above issues. For our initial prototype, Freeze Inference creates an offline cache, which stores the intermediate layer outputs computed over training data. We then train a dimensionality reduction model for each layer, which projects high dimensional intermediate layer tensors to a low dimensional space. Finally, we perform $k$-means clustering on the reduced dimensional space and only store the centroids of the clusters, which further reduces the memory footprint and look-up time.             

The rest of the paper is organized as follows. In Sec. \ref{sec:Intuition}, we introduce the rationale behind caching in the context of DNNs. Next, we describe the design of our system Freeze Inference in Sec. \ref{sec:design}. Finally, we present the initial results of our prototype (Sec.~\ref{sec:res}) and conclude with discussions on future research directions (Sec.~\ref{sec:disc}).

\section{Intuition}\label{sec:Intuition}
We begin by describing some key properties of DNNs and how layer-wise caching as we envision it can be applied in this setting. In a DNN, the input is represented as a set of features, where each feature is a value provided to individual nodes at the DNN's input layer. The DNN has a number of hidden layers each consisting of multiple nodes, where each node applies a non-linear function to a weighted sum of its inputs. We refer to the individual hidden layer outputs as {\it intermediate layer} outputs in our work. Finally, there is an output layer which consists of one or more neurons that can cumulatively be viewed as making a prediction.

Given the architecture of DNNs, we make two important observations which form the basis for our work:
    
    \noindent {\bf (O1)} Given two inputs $X_{i}$ and $X_{j}$ which are exactly same, the DNN will predict the same label $Y$ for both the inputs. This is because the same set of learned weights are used during inference which effectively means that each layer executes a deterministic function on its input. On similar lines, if two inputs $X_{i}$ and $X_{j}$ result in the same intermediate layer output at a given hidden layer, we expect the DNN to predict the same label for both the inputs.
    
    \noindent {\bf (O2)} Consider two inputs $X_{i}$ and $X_{j}$ whose output feature vectors reside close to each other in the output feature space. To predict labels, DNNs typically use a function such as softmax at the output layer which draws decision boundaries in the output feature space. Proximity in the output feature space means there is a high probability that the points lie within the same decision boundary and are assigned the same prediction $Y$ by the DNN.

\begin{figure}[t]
  \centering
   \includegraphics[width=0.4\textwidth]{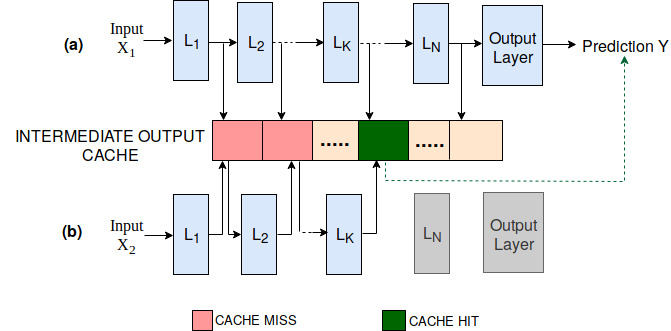}
    \caption{The basic idea behind Freeze Inference. (a) The intermediate outputs are cached. (b) During inference, a cache look-up is done after every layer and a cache hit yields a faster prediction}
    \label{fig:Intuition}
\end{figure}

Figure \ref{fig:Intuition} builds up towards the intuition behind Freeze Inference. Consider a DNN with $N$ layers. Let us take two input feature vectors $X_{1}$ and $X_{2}$. Let us say that each input produces intermediate layer outputs $L_{i,j}$, where $i$ is the layer number and $j$ is an identifier for the input under consideration. Now, let us consider a situation where $X_{1}$ has already run through the DNN to obtain a prediction $Y$. We store each of it's intermediate layer outputs $L_{i,1}$ for $i=1,2,3..N$ along with the final predicted label $Y$ in a per-layer cache. We now try to predict the label for input $X_{2}$  as follows: after the computation at each layer, we additionally compare the obtained intermediate output to the contents of the corresponding layer's cache. During this process, let us say that at some layer $K$ we observe that $L_{K,1}$ and $L_{K,2}$ are the same. From observation O1, we can conclude that the intermediate outputs of successive layers would also be the same ultimately leading to the same prediction. More formally, if $K$ is the smallest layer at which we have $L_{K,1}$ = $L_{K,2}$, then we can say that $L_{i,1}$ = $L_{i,2}$ for $i=K+1,K+2,..N$ and both $X_{1}$ and $X_{2}$ would have the same predicted label $Y$. Hence, it is possible to skip expensive computations for layer $K+1,K+2,..N$ when the intermediate layer output for a layer $K$ matches an intermediate output that has been cached. In such a scenario, we can \textit{freeze} the computation at layer $k$ and return the cached output.
\subsection{Towards Approximate Caching} \label{subsec:approx_caching}
\begin{figure}[t!]
  \centering
    \includegraphics[width=0.35\textwidth]{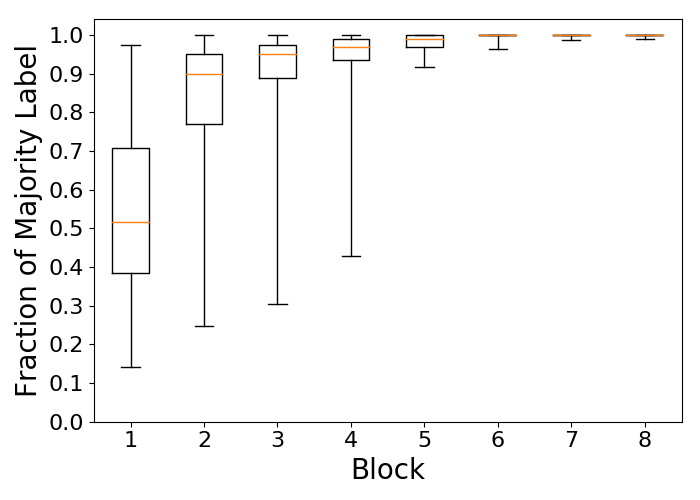}
    \vspace{-10pt}
    \caption{Distribution of fractional share of majority label per cluster for ResNet-18 on CIFAR-10}
    \label{fig:clustering}
\end{figure}

Since feature vectors have high dimensionality and are represented by a set of floats, it is highly unlikely that that two intermediate layer outputs would be exactly the same. Therefore, a caching mechanism based on exact matches would not generate enough cache hits to provide meaningful computational benefits. To address this, we leverage the insight from observation O2 in that DNNs try to identify decision boundaries in order to classify items. To empirically validate this, we took a set of 50,000 images belonging to the CIFAR-10\cite{cifar_10} dataset and ran a complete forward pass for each image on the ResNet-18\cite{Resnet} model. From this, we constructed a set of intermediate layer outputs for each ResNet block\footnote{ResNet-18 consists of 8 blocks, where each block consists of 2 convolutional layers and 1 residual connection} and tagged each output with the label predicted by the model. For each ResNet block, we then arranged the outputs into 200 clusters using $k$-means and computed the majority label occupying each cluster along with the fractional share of the label within that cluster. From Figure \ref{fig:clustering}, we notice that there is a dominant majority label in each cluster. For instance, in Block 4, we see that the mean fraction of the majority label is 0.95. This provides empirical backing that there exists a semantic relationship between points that lie nearby to each other in the intermediate feature space. Another interesting observation is that the mean fraction of the majority label increases as we move across the layers, indicating that points get better correlated in the feature space as we go deeper in the DNN.

We leverage the above ideas in Freeze Inference by constructing an offline, per-layer cache consisting of intermediate layer outputs and their corresponding labels. We augment the inference control flow to perform an approximate cache look-up for the intermediate output at each layer by using an algorithm like $k$-nearest neighbors. We use information such as the labels of the $k$ neighbors and their distances from the input point to make a prediction and offer a notion of {\it confidence} about the prediction. We characterize the cache look-up at that layer as a {\it hit} if the offered {\it confidence} exceeds a defined $threshold$ for that layer.


\section{Freeze Inference Design}\label{sec:design}
The first major challenge in Freeze Inference is that the intermediate layer outputs reside in high-dimensional space. Apart from resulting in low cache hit rates, high memory usage, and increasing the computational complexity of cache look-up, prior work\cite{Weber:1998:QAP:645924.671192} has shown that performance of similarity search degrades in high dimension. This is a problem since Freeze Inference relies on the notion of closeness to infer semantic similarity. We overcome this by using dimensionality reduction. Inspired by Metric Learning~\cite{xing2003distance}, we do this using a one layer neural network whose hidden layer consists of 1024 nodes. We train a per-layer dimensionality reduction model using the intermediate layer outputs and labels predicted by the model for the training data set. 

Next, we define the semantics of approximate cache look-ups by describing a cache look-up API. The API takes in an intermediate layer representation as an argument and returns a prediction along with a confidence value. Following from the discussion in Section \ref{subsec:approx_caching}, the API computes $k$-nearest neighbors on the input and obtains a set of $k$ tuples, where each tuple consists of the label of the neighbor and its distance from the input. In our initial design, we use a heuristic for computing the predicted label and the associated confidence value. Our heuristic is based on the intuition that predictions can be more confident if (a) more neighbors agree on the same label and (b) the neighbors are close to the input under consideration. Let us say that the dataset has $N$ labels $n_{1}, n_{2}...,n_{N}$. For a given input point, suppose label $n_{i}$ has $m_{i}$ occurrences amongst the $k$ neighbors for the input at a specific layer of the DNN. Let the distances associated with the $m_{i}$ occurrences be $d_{1},.,d_{j},.,d_{m_{i}}$. We first compute $n_{i}$ label's fractional share amongst the $k$ neighbors as $S_{i}=\frac{m_{i}}{k}$. We then compute the confidence $C_{i}=S_{i}\times\sum_{j=1}^{m_{i}}{\frac{1}{d_{j}}}$ for each label. The API returns the label having maximum confidence as the predicted label for the layer along with the associated confidence.

\begin{figure}[t!]
  \centering
    \includegraphics[width=0.5\textwidth]{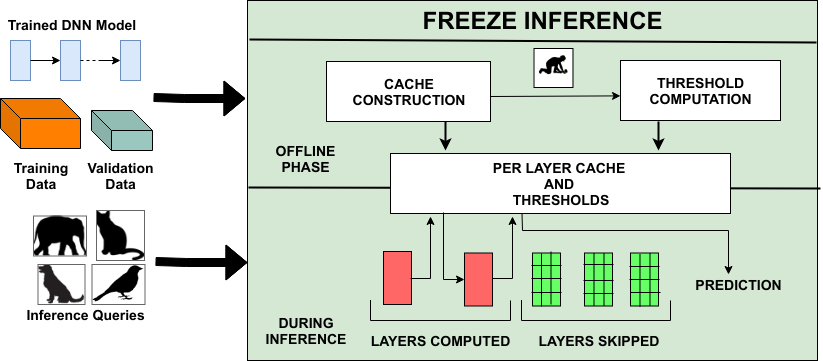}
    \caption{Freeze Inference High Level Design}
    \label{fig:freeze_hld}
\end{figure}

Figure \ref{fig:freeze_hld} presents the Freeze Inference pipeline. It consists of an offline phase which aggregates information to be used during inference. The offline phase consists of two parts - (i) Cache Construction  (ii) Threshold Computation. The above two steps lead to the construction of per-layer caches and per-layer thresholds which are then passed onto the online phase. These structures are used during the online inference phase to perform cache look-up and characterize cache hits. We describe the individual steps in detail below.

\label{sec:design}

\floatname{algorithm}{Pseudocode}
\begin{algorithm}[t!]
\begin{scriptsize}
\begin{algorithmic}[1]
\State \textsc{cache} = \{\} \Comment{Per-layer cache}
\State \textsc{thresholds} = \{\} \Comment{Per-layer thresholds}
\State{}
\State {$\triangleright$ Given a model M, perform offline pre-processing for Freeze Inference}
\Procedure{OfflinePhase}{Model $M$, TrainData $TD$, ValidationData $VD$}
\label{offline_phase}
       \State \textsc{cache} = \textsc{ConstructCache}($M$, $T$)
       \State \textsc{thresholds} = \textsc{ComputeThresholds}($M$, $V$, \textsc{cache})
\EndProcedure
\State{}

\State {$\triangleright$ Given a model $M$, perform Freeze Inference on input $I$}
\Procedure{FreezeInference}{Model $M$, Input $I$}
\label{start_hld_inference}
       \ForAll{$layer \in M.layers()$}
                \State $layerOutput$ = forward pass on $M$ for next $layer$
                \State $pred\_label$, $confidence$ = prediction for $layerOutput$ from \textsc{cache}[$layer$]
                \If{$confidence$ > \textsc{thresholds}[$layer$]}
                       \State return $predicted\_label$
                \EndIf
       \EndFor
       \State return $label$ predicted by output layer of $M$
\EndProcedure \label{end_hld_freeze}
\State{}

\Procedure{ConstructCache}{Model $M$, TrainData $TD$} \label{start_construct_cache}
    \State $IO[i][j]$ \Comment{Intermediate output for $TD$[$i$] at layer $j$}
    \State $Y[i]$ \Comment{Label Predicted by $M$ for $TD$[$i$]}
    \ForAll{$item \in TD$}
        \ForAll{$layer \in M.layers()$}
            \State \textsc{cache}[$layer$].append(<$IO$[$item$][$layer$], $Y$[$item$]>)
        \EndFor
    \EndFor
\EndProcedure \label{end_construct_cache}
\State{}

\Procedure{ComputeThresholds}{Model $M$, ValidationData $VD$, Cache $C$}
    \State $IO[i][j]$ \Comment{Intermediate output for $VD$[$i$] at layer $j$}
    \State $Y[i]$ \Comment{Label Predicted by $M$ for $VD$[$i$]}
    \State $Prediction[i][j]$ \Comment{Label Predicted by look-up from $C$ for $VD$[$i$] at layer $j$}
    \State $Confidence[i][j]$ \Comment{Confidence value of look-up from $C$ for $VD$[$i$] at layer $j$}
    \ForAll{$item \in VD$}
        \ForAll{$l \in M.layers()$}
            \If{$Prediction$[$item$][$l$] {\bf not equals} Y[$item$]}
            \State \textsc{threshold}[$l$] = $max$(\textsc{threshold}[$l$], $Confidence$[$item$][$l$]) \label{threshold_logic}
            \EndIf
        \EndFor
    \EndFor
\EndProcedure
\Statex
\end{algorithmic}
\end{scriptsize}
\caption{Freeze Inference Workflow}
\label{alg:freeze_inference}
\vspace{-0.3cm}
\end{algorithm}

\subsection{Offline Phase}

\noindent {\bf (i) Cache Construction:}
    Freeze Inference takes in a trained model and constructs per-layer caches by running a forward pass of the DNN for each example and caching the dimensionally reduced intermediate layer outputs for each layer (Line \ref{start_construct_cache}--\ref{end_construct_cache} in Algorithm \ref{alg:freeze_inference}).    
    
    \noindent {\bf (ii) Threshold Computation:} 
    A critical piece of the Freeze Inference design is to develop the notion of a cache hit. For this purpose, we use a validation dataset to compute per-layer thresholds. For each item in the validation data, we perform a forward pass, reduce the dimension of the layer output and then do a cache look-up at each layer. For each layer, we set the threshold as the maximum confidence value that resulted in a wrong prediction on the validation set (Line \ref{threshold_logic} in Algorithm \ref{alg:freeze_inference}). Thus, Freeze Inference adopts a pessimistic approach by establishing strict thresholds and ensuring zero error on the validation data, which in turn maximizes the accuracy of cache hits during inference.
    
\vspace{-0.4cm}
\subsection{Online Phase - Inference}

When an inference request comes in, we do forward propagation one layer at a time and a cache look-up on the dimensionally reduced output at each layer. If the confidence returned by cache look-up is greater than the established threshold for that layer, we skip the computation of the remaining layers and return the label predicted by cache look-up as the final predicted label. (Lines \ref{start_hld_inference} to \ref{end_hld_freeze} in Algorithm \ref{alg:freeze_inference}).

\section{Results and Challenges}\label{sec:res}
\begin{figure}
\centering
    \subfloat[Upper Bound]{\includegraphics[width=0.49\columnwidth]{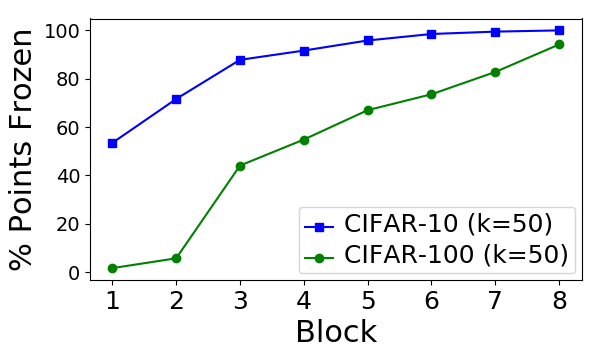}} 
    \subfloat[Actual]{\includegraphics[width=0.49\columnwidth]{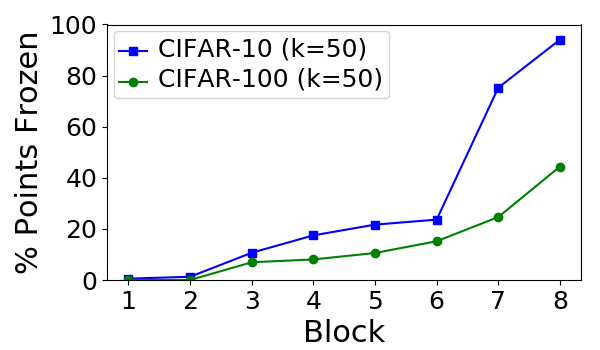}} 
    \vspace{-8pt}
\caption{\label{fig:knn}
\footnotesize
CDF of points frozen w.r.t. blocks for ResNet-18 using $k$-NN
}

\end{figure}
We evaluate our Freeze Inference system for the CIFAR-10\cite{cifar_10} and CIFAR-100\cite{cifar_100} datasets on the ResNet-18 model\cite{Resnet}. 
Figure \ref{fig:knn}(a) presents the earliest block at which an inference request can be $frozen$ assuming that we have a perfect threshold calculation scheme. In this scenario, we observe that can potentially save half the computation time (run half of the total layers) for around 91\% data-points in CIFAR-10  and 55\% of the data-points for CIFAR-100. This represents an upper bound on the potential of Freeze Inference. Figure \ref{fig:knn}(b) shows the distribution of layers at which inference requests are $frozen$ using our naive threshold calculation scheme. From the graph, we observe that our naive Freeze Inference saves half the computation for about 20\% of the points on CIFAR-10 and around 15\% for CIFAR-100 respectively. Overall, we are able to freeze about 95\% and 44\% of the data-points before the output layer in CIFAR-10 and CIFAR-100 respectively. The $k$-NN approach with our naive threshold calculation scheme achieves an accuracy of 97.48\% and 99.03\% with respect to the model's prediction for CIFAR-10 and CIFAR-100 datasets respectively. This shows that we are able to save computation without trading off too much on the accuracy.

\begin{table}[t]
\vspace{-0.3cm}
  \centering
  \begin{tabular}{ l | r }
    \hline
    \textbf{Cache Construction Scheme} & \textbf{Total Memory} \\
    \hline
    k-NN without Dimensionality Reduction & 37500.0 MB \\
    \hline
    k-NN with Dimensionality Reduction & 2500.0 MB \\
    k-Means with Dimensionality Reduction & 12.5 MB \\
    \hline
  \end{tabular}
  \caption{Memory requirements for caching on ResNet-18}
  \label{tab:memory_reqs}
\end{table}

Even though we are able to Freeze a significant percentage of data-points, $k$-NN has following overheads:

    \noindent {\bf Computational Complexity: } Computing $k$-nearest neighbors incurs significant overheads due to a large number of cached intermediate points to compare against. In our experiment, we had 40,000 training data points in the cache at each layer. To extract the most out of Freeze Inference, we would require the cache-lookup to be computationally cheap.
    \noindent {\bf Memory Overheads: } With caching, Freeze Inference incurs an additional overhead with respect to memory. Table \ref{tab:memory_reqs} captures memory usage that the $k$-NN implementation of Freeze Inference would require. Though dimensionality reduction significantly reduces the memory overhead, we would still like the requirement to be as low as possible to allow Freeze Inference to scale well for larger datasets and models.

We can overcome the computational and memory overheads by leveraging the fact that intermediate layer points are semantically related to each other (Figure \ref{fig:clustering}). In this light, we use $k$-means to cluster neighboring intermediate layer points and represent them by a single cluster center. This reduces the number of points that need to be stored in the cache and consequently reduces both the computational complexity and memory overheads. We construct a per-layer cache such that each item consists of a cluster center, the majority label and the fraction of majority label in that cluster. During inference, the API returns the majority label of the closest cluster as the predicted label and the ratio of the fraction of majority label to distance from the cluster center as the confidence value. The thresholding scheme used by $k$-means is the similar to the one described earlier for $k$-NN.

\begin{figure}
\centering
    \subfloat[ResNet-18]{\includegraphics[width=0.49\columnwidth]{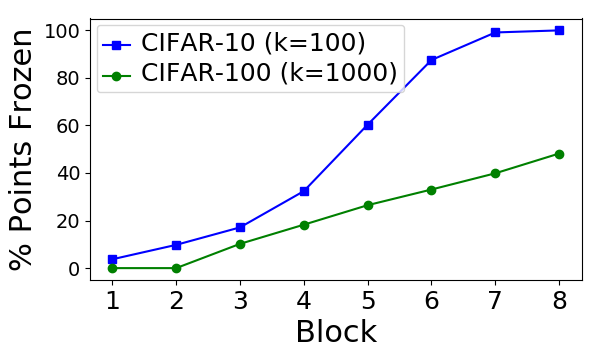}} \label{fig:upper_bound}
    \subfloat[ResNet-50]{\includegraphics[width=0.49\columnwidth]{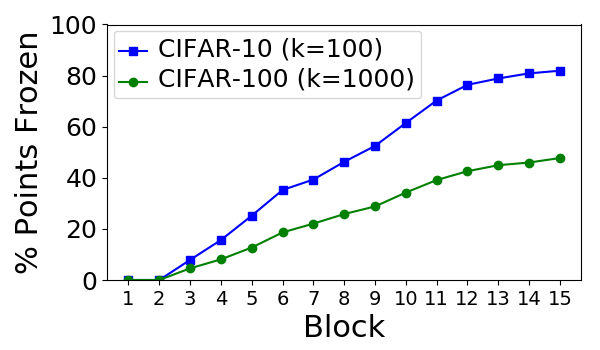}} \label{fig:knn_res18}
\caption{\label{fig:kmeans}
\footnotesize
CDF of points frozen w.r.t. blocks using $k$-means
}   
\end{figure}
\begin{figure}
\centering
    \subfloat[Block 7]{\includegraphics[width=0.49\columnwidth]{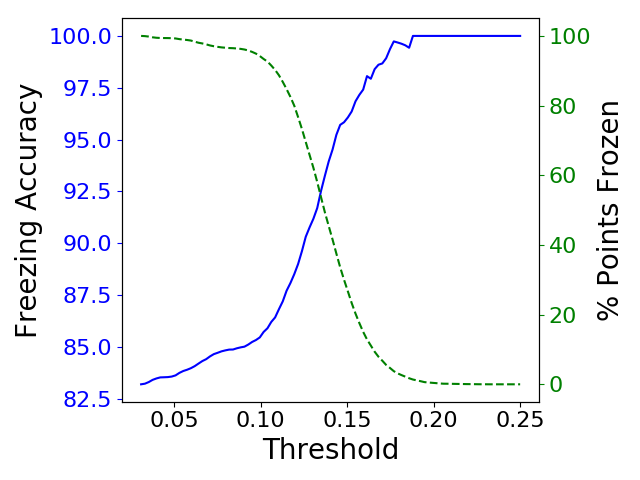}}
    \subfloat[Block 15]{\includegraphics[width=0.49\columnwidth]{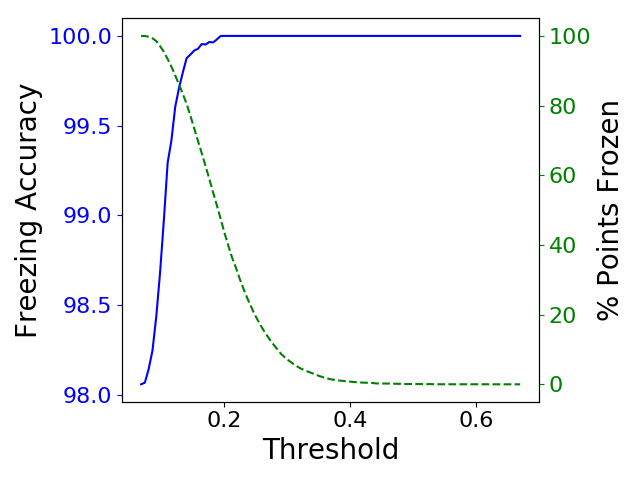}}
\vspace{-10pt}
\caption{\label{fig:tradeoff}
\footnotesize
Trade-off between accuracy of frozen points and percentage of total points frozen as the threshold varies
for block 7 and block 15 of ResNet-50}
\end{figure}

Figure \ref{fig:kmeans} shows the distribution of layers at which inference requests are $frozen$ using the $k$-means clustering approach for ResNet-18 and ResNet-50\footnote{ResNet-50 consists of 15 blocks, where each block consists of 3 convolutional layers and 1 residual connection.}. For ResNet 18, we observe that Freeze Inference saves half the computation for about 33\% and 19\% of the points on CIFAR-10 and CIFAR-100 respectively. Similarly, for ResNet-50, we are able to save half the computation for 46\% and 26\% of the points on CIFAR-10 and CIFAR-100 respectively. Overall, this approach achieves an accuracy of 92.85\% and 88.86\% with respect to the model's prediction for CIFAR-10 and CIFAR-100 datasets respectively. Figure \ref{fig:tradeoff} captures the trade-off between the percentage of points that Freeze Inference can $freeze$ and the accuracy of those points for CIFAR-10 on ResNet-50. We observe that as we increase the threshold, the percentage of points frozen decreases which results in points getting frozen more accurately. This tells us that we can model thresholding as an optimization problem where we need to simultaneously maximize the percentage of points frozen and the accuracy with which they are frozen.

With respect to computation, our experiments show that cache look-up is about 20$X$ faster than the compute for a single layer. Additionally, from Table \ref{tab:memory_reqs} we see that the cache requires a mere 12.5MB of memory for ResNet-18. Thus, $k$-means solves the problems both with respect to computational complexity and memory requirements.

\section{Research Directions}\label{sec:disc}
\label{sec:discussion}

\noindent{\bf Cache Size and Accuracy Trade-off: } From the results, we notice that while $k$-means is able to lower the memory requirements this comes at the cost of reduced accuracy. We plan to study techniques that can further improve the heuristic used to compute confidence and thresholds per layer and also study the effect of disabling freeze inference at earlier layers (e.g., Block 4) as most of errors happen in the initial few layers.

 
\noindent{\bf Freeze Inference on GPUs: } Batching of inference requests is a popular technique used to increase prediction throughput. With Freeze Inference, we would need to reconstruct the batch after each layer to remove the items that have been \emph{frozen}. Handling dynamic batch sizes in GPUs across layers is an interesting research problem that needs to be investigated.

\noindent{\bf Cache lookup performance:} Optimizing the cache lookup performance is important for realizing the benefits from freezing. While our current prototype uses a single thread on CPU to compute distance from centroids, we plan to investigate techniques to pipeline cache lookups with the forward pass being executed on GPUs.

    
\noindent{\bf Incremental Cache Update: } In the current design, we use a cache that is constructed offline from the training data. To handle updates, we plan to identify common inference requests that were not frozen over a period of time, collect their intermediate layer representations and labels and once enough examples have accumulated, we can use these examples to recompute the thresholds. Performing online cache updates is a challenging problem especially with $k$-means as clusters and thresholds need to be re-computed for every update.

\vspace{0.2cm}

\noindent {\bf Acknowledgements.} We thank Yingyu Liang, Arjun Singhvi and reviewers for their valuable feedback and suggestions. This work is supported by the National Science Foundation (CNS-1838733). Shivaram Venkataraman is also supported by a Facebook faculty research award and support for this research was also provided by the Office of the Vice Chancellor for Research and Graduate Education at the University of Wisconsin, Madison with funding from the Wisconsin Alumni Research Foundation. Aditya Akella is also supported by a Google Faculty award, a gift from Huawei, and H. I. Romnes Faculty Fellowship.



\section{Discussion Topics}
In this paper we introduced Freeze Inference, a general technique to improve the latency of serving deep learning models by using cache and have shown that this direction has potential. This paper is likely to generate a discussion regarding the opportunities for not running all the layers of a DNN. Some points that we think will lead to discussion include:

\noindent\textbf{Design approaches for an approximate cache}: In this paper, we presented an initial approach at designing an approximate cache using $k$-NN and $k$-means clustering. If other techniques can improve the trade-off between cache size, cache lookup time, and accuracy, it will make for an interesting discussion.

\noindent\textbf{Dynamic batching on GPUs}: As discussed in Section~\ref{sec:disc}, the problem of dynamically adjusting the batch size on GPUs is very interesting from a systems perspective. Solutions to this problem could also improve other techniques like SkipNets~\cite{DBLP:journals/corr/abs-1711-09485}.

\noindent\textbf{Effect of non-uniform request popularity}: Finally our evaluation results consider a uniform distribution of requests from the test dataset. However in real world scenarios we often see a zipfian pattern with a few very popular requests and it will be interesting to discuss on how we can achieve greater benefits for such requests. \\

{
\bibliographystyle{acm}
\bibliography{references} 

\begin{thebibliography}{10}

\bibitem{DBLP:journals/corr/CaiHSV17}
{\sc Cai, Z., He, X., Sun, J., and Vasconcelos, N.}
\newblock Deep learning with low precision by half-wave gaussian quantization.
\newblock {\em CoRR abs/1702.00953\/} (2017).

\bibitem{DBLP:journals/corr/CourbariauxB16}
{\sc Courbariaux, M., and Bengio, Y.}
\newblock Binarynet: Training deep neural networks with weights and activations
  constrained to +1 or -1.
\newblock {\em CoRR abs/1602.02830\/} (2016).

\bibitem{crankshaw2017clipper}
{\sc Crankshaw, D., Wang, X., Zhou, G., Franklin, M.~J., Gonzalez, J.~E., and
  Stoica, I.}
\newblock Clipper: A low-latency online prediction serving system.
\newblock In {\em 14th {USENIX} Symposium on Networked Systems Design and
  Implementation ({NSDI} 17)\/} (Boston, MA, 2017), {USENIX} Association,
  pp.~613--627.

\bibitem{hassan2018achieving}
{\sc Hassan, H., Aue, A., Chen, C., Chowdhary, V., Clark, J., Federmann, C.,
  Huang, X., Junczys-Dowmunt, M., Lewis, W., Li, M., et~al.}
\newblock Achieving human parity on automatic chinese to english news
  translation.
\newblock {\em arXiv preprint arXiv:1803.05567\/} (2018).

\bibitem{Resnet}
{\sc He, K., Zhang, X., Ren, S., and Sun, J.}
\newblock Deep residual learning for image recognition.
\newblock {\em CoRR abs/1512.03385\/} (2015).

\bibitem{hinton2015distilling}
{\sc Hinton, G., Vinyals, O., and Dean, J.}
\newblock Distilling the knowledge in a neural network.
\newblock {\em arXiv preprint arXiv:1503.02531\/} (2015).

\bibitem{DBLP:journals/corr/HuangCLWMW17}
{\sc Huang, G., Chen, D., Li, T., Wu, F., van~der Maaten, L., and Weinberger,
  K.~Q.}
\newblock Multi-scale dense convolutional networks for efficient prediction.
\newblock {\em CoRR abs/1703.09844\/} (2017).

\bibitem{cifar_10}
{\sc Krizhevsky, A., Nair, V., and Hinton, G.}
\newblock Cifar-10 (canadian institute for advanced research).

\bibitem{cifar_100}
{\sc Krizhevsky, A., Nair, V., and Hinton, G.}
\newblock Cifar-100 (canadian institute for advanced research).

\bibitem{NIPS2012_4824}
{\sc Krizhevsky, A., Sutskever, I., and Hinton, G.~E.}
\newblock Imagenet classification with deep convolutional neural networks.
\newblock In {\em Advances in Neural Information Processing Systems}. 2012,
  pp.~1097--1105.

\bibitem{pretzel}
{\sc Lee, Y., Scolari, A., Chun, B.-G., Santambrogio, M.~D., Weimer, M., and
  Interlandi, M.}
\newblock {PRETZEL}: Opening the black box of machine learning prediction
  serving systems.
\newblock 611--626.

\bibitem{DBLP:journals/corr/LinCY13}
{\sc Lin, M., Chen, Q., and Yan, S.}
\newblock Network in network.
\newblock {\em CoRR abs/1312.4400\/} (2013).

\bibitem{mcdnnPaper}
{\sc Shen, H., Philipose, M., Agarwal, S., and Wolman, A.}
\newblock Mcdnn: An execution framework for deep neural networks on
  resource-constrained devices.
\newblock Tech. rep., December 2015.

\bibitem{DBLP:journals/corr/SimonyanZ14a}
{\sc Simonyan, K., and Zisserman, A.}
\newblock Very deep convolutional networks for large-scale image recognition.
\newblock {\em CoRR abs/1409.1556\/} (2014).

\bibitem{43022}
{\sc Szegedy, C., Liu, W., Jia, Y., Sermanet, P., Reed, S., Anguelov, D.,
  Erhan, D., Vanhoucke, V., and Rabinovich, A.}
\newblock Going deeper with convolutions.
\newblock In {\em Computer Vision and Pattern Recognition (CVPR)\/} (2015).

\bibitem{DBLP:journals/corr/SzegedyVISW15}
{\sc Szegedy, C., Vanhoucke, V., Ioffe, S., Shlens, J., and Wojna, Z.}
\newblock Rethinking the inception architecture for computer vision.
\newblock {\em CoRR abs/1512.00567\/} (2015).

\bibitem{DBLP:journals/corr/abs-1711-09485}
{\sc Wang, X., Yu, F., Dou, Z., and Gonzalez, J.~E.}
\newblock Skipnet: Learning dynamic routing in convolutional networks.
\newblock {\em CoRR abs/1711.09485\/} (2017).

\bibitem{Weber:1998:QAP:645924.671192}
{\sc Weber, R., Schek, H.-J., and Blott, S.}
\newblock A quantitative analysis and performance study for similarity-search
  methods in high-dimensional spaces.
\newblock In {\em VLDB\/} (1998), pp.~194--205.

\bibitem{xing2003distance}
{\sc Xing, E.~P., Jordan, M.~I., Russell, S.~J., and Ng, A.~Y.}
\newblock Distance metric learning with application to clustering with
  side-information.
\newblock In {\em Advances in neural information processing systems\/} (2003),
  pp.~521--528.

\bibitem{toward-human-parity-conversational-speech-recognition}
{\sc Xiong, W., , Huang, X., Seide, F., , and Stolcke, A.}
\newblock Toward human parity in conversational speech recognition.
\newblock {\em IEEE/ACM Transactions on Audio, Speech, and Language Processing
  25\/} (Sept 2017), 2410--2423.

\end{thebibliography}
}
\end{document}